\documentclass[11pt,a4paper]{article}
\usepackage[hyperref]{acl2021}
\usepackage{times}
\usepackage{latexsym}
\usepackage{float}
\usepackage{lipsum}
\usepackage{amsfonts}
\usepackage{amsmath}
\usepackage{graphicx}

\usepackage{microtype}

\aclfinalcopy 

\title{Are Pre-trained Convolutions Better than Pre-trained Transformers?}
\author{Yi Tay \\
  Google Research\\
  Mountain View, California\\
  \texttt{yitay@google.com} \\
  \And
  Mostafa Dehghani \\ 
   Google Research, Brain Team\\
   Amsterdam, Netherlands \\ 
   \texttt{dehghani@google.com}
   \AND
   Jai Gupta \\ 
   Google Research \\ 
  Mountain View, California \\ 
   \texttt{jaigupta@google.com} \\
   \And 
   Vamsi Aribandi\thanks{\hspace{1.5mm}Google AI Resident} \\
   Google Research \\
  Mountain View, California \\ 
   \texttt{aribandi@google.com}
   \And
   Dara Bahri \\ 
   Google Research \\ 
   Mountain View, California\\
   \texttt{dbahri@google.com}
   \AND
   Zhen Qin \\ 
   Google Research \\
  Mountain View, California \\
   \texttt{zhenqin@google.com}
   \And 
   Donald Metzler \\ 
   Google Research \\ 
   Mountain View, California \\
    \texttt{metzler@google.com}
  }
\date{}

\begin{document}
\maketitle
\begin{abstract}
In the era of pre-trained language models, Transformers are the de facto choice of model architectures. While recent research has shown promise in entirely convolutional, or CNN, architectures, they have not been explored using the pre-train-fine-tune paradigm. In the context of language models, are convolutional models competitive to Transformers when pre-trained? This paper investigates this research question and presents several interesting findings. Across an extensive set of experiments on 8 datasets/tasks, we find that CNN-based pre-trained models are competitive and outperform their Transformer counterpart in certain scenarios, albeit with caveats. Overall, the findings outlined in this paper suggest that conflating pre-training and architectural advances is misguided and that both advances should be considered independently. We believe our research paves the way for a healthy amount of optimism in alternative architectures.
\end{abstract}

\section{Introduction}
In the modern era of pre-training, there appears to be an unbreakable tie between Transformer architectures \cite{vaswani2017attention} and pre-trained language models. Models such as BERT \cite{devlin2018bert}, RoBERTa \cite{liu2019roberta}, and T5 \cite{raffel2019exploring} have all adopted Transformers as their underlying architecture. As a matter of fact, there are barely any recent pre-trained models not based on Transformers.

While the contextual representation learning has a rich history \citep{pennington2014glove,dai2015semi,chidambaram2018learning,liu2020survey,qiu2020pre}, modern pre-trained language modeling started with models like ELMo \cite{peters2018deep} and CoVE \cite{mccann2017learned} which are based on recurrent (e.g. LSTM \citep{hochreiter1997long}) architectures. Although they were successful, research using these architectures dwindled as Transformers stole the hearts of the NLP community, having, possibly implicitly, been perceived as a unequivocal advancement over its predecessors.

Recent work demonstrates the promise of entirely convolution-based models \cite{wu2019pay,gehring2017convolutional} and questions the necessity of self-attentive architectures like Transformers. For example, in \cite{wu2019pay}, the proposed convolutional seq2seq models outperform Transformers on a series of canonical benchmarks such as machine translation and language modeling. From these findings emerge a rather natural line of questioning - should we consider pre-trained models beyond Transformers?

Despite early success, the relevance of convolutional models in the era of pre-trained language models remains an open question. To the best of our knowledge, convolutional architectures have not yet been rigorously evaluated under the \emph{pre-train-fine-tune} paradigm. This is the primary purpose of this work.
Concretely, this paper seeks to empirically validate whether pre-trained convolutions are competitive with pre-trained Transformers across a range of tasks. 

The interaction between pre-training schemes and model architectures is an under-studied topic. Are only Transformers able to capitalize on the benefits of pre-training? If we use a different architectural inductive bias, would there also be a substantial gain unlocked by pre-training? Are pre-trained convolutions better in particular scenarios? This paper investigates these questions.

There are a number of obvious benefits of convolution-based models. Firstly, convolutions do not suffer from the quadratic memory complexity of self-attention - a problem significant enough that it spawned the creation of the entirely new category of ``efficient'' Transformer architectures \cite{tay2020efficient,tay2021long}. Secondly, convolutions operate locally and do not rely on positional encodings as an order signal to the model. That said, convolutions also come with a slew of downsides. For example, being unable to access global information means such models are unable to perform a form of cross-attention across multiple sequences. We dive into the details of this more in subsequent sections. 

In this paper, we present a pre-trained convolutional sequence-to-sequence, or Seq2Seq, model. We train our convolutional model using span-based sequence-to-sequence denoising objectives similar to those employed in T5 \cite{raffel2019exploring}. We evaluate a variety of convolutional variants (e.g., dilated, lightweight, dynamic \cite{wu2019pay}, etc.) under both raw (no pre-training) and pre-train-fine-tune paradigms. Our goal is to understand the true competitiveness of convolutional architectures in the era of pre-training. 

We show that pre-trained convolutions are competitive against pre-trained Transformers via a set of experiments on a potpourri of NLP tasks, like toxicity detection, sentiment classification, news classification, query understanding and semantic parsing/compositional generalization \cite{kim2020cogs}.  Moreover, we find that pre-trained convolutions can outperform, in terms of model quality and training speed, state-of-the-art pre-trained Transformers \citep{raffel2019exploring} in certain scenarios. However, to provide a balanced perspective, we also describe scenarios where pre-trained convolutions do not perform well and may be deemed unsuitable.
\paragraph{Contributions}
Overall, the main contributions of this paper can be summarized as follows:
\begin{itemize}
    \item We perform a comprehensive empirical evaluation of convolutional Seq2Seq models under the \textit{pre-train-fine-tune} paradigm. To the best of our knowledge, the competitiveness and relevance of pre-trained convolutions still remains an open question. 
    \item We make several important observations. Specifically, we find that (1) pre-training helps convolutional models just as much as it helps Transformers, and (2) pre-trained convolutions are competitive alternatives in certain scenarios in terms of model quality and training speed. 
    \item We conduct extensive experiments across 8 datasets spanning a diverse range of tasks and domains. On 7 out of 8 tasks, we find that pre-trained convolutions outperform a recent state-of-the-art transformer (T5 \citep{raffel2019exploring}) with and without pre-training. We examine the speed and operation count (FLOPS) of convolutions versus Transformers and find that convolutions are not only faster but also scale better to longer sequence lengths.
    \item Our checkpoints and code are available here\footnote{ \url{https://github.com/google-research/google-research/tree/master/pretrained_conv}}. Notably, the core model functionality can be found in Mesh Tensorflow\footnote{\url{https://github.com/tensorflow/mesh}}. If permissions are a problem please check this cloud bucket\footnote{\url{gs://scenic-bucket/pretrainedconvs/pretrainedconvs}} for the checkpoints.

\end{itemize}

\section{Related Work}
Pre-training on a large corpus has become the primary method of learning universal language representations to solve different downstream NLP tasks. 
The first generation of pre-trained models aimed at learning embedding for words, like Skip-Gram~\citep{mikolov2013distributed} and Glove~\citep{pennington2014glove}, and quickly developed to learning contextualized representation for words, like ELMO~\citep{peters2018deep}, GPT~\citep{radford2018improving}, and BERT~\citep{devlin2018bert}. This, however, is not the only axis in which pre-trained models have evolved. 

Different objective functions and various tasks, both supervised and unsupervised, have been explored for pre-training. For instance, CoVe~\citep{mccann2017learned} uses machine translation as the pre-training task, ELMO~\citep{peters2018deep} and GPT~\citep{radford2018improving} use language modeling objectives, BERT~\citep{devlin2018bert} uses masked language modeling, T5~\citep{raffel2019exploring} and MASS~\citep{song2019mass} use Seq2Seq masked language modeling, and XLNet~\citep{yang2019xlnet} utilizes permuted language modeling. In addition to this, BART~\citep{lewis2019bart} uses a denoising autoencoder setup during pre-training, where the model takes a partially corrupted input and is trained to recover the original, undistorted input. Some models use a contrastive learning setup during pertaining, like replaced token detection, used by ELECTRA~\citep{clark2020electra}, and sentence order prediction, used by ALBERT~\citep{lan2019albert} and StructBERT~\citep{wang2019structbert}.

Another axis where pre-trained models in NLP explored different ideas is model architecture. ELMO~\citep{peters2018deep} and CoVe~\citep{mccann2017learned} used LSTMs as the base model. Later, Transformers~\citep{vaswani2017attention} became the de facto architecture of pre-trained NLP models. BERT~\citep{devlin2018bert}, XLNet~\citep{yang2019xlnet} and RoBERTa~\citep{liu2019roberta} use the Transformer encoder, while GPT~\citep{radford2018improving}, GPT-2~\citep{radford2019language}, and GPT-3~\citep{brown2020language} use the Transformer decoder as the backbone. Some pre-trained models are also are based on the encoder-decoder transformer architecture, like T5~\citep{raffel2019exploring}, MASS~\citep{song2019mass}, and BART~\citep{lewis2019bart}. In this paper, we investigate another model architecture variation by studying the power of convolutional neural network as the backbone of pre-trained models for NLP.

Convolutions have always been an interesting choice for sequence modeling and NLP applications \citep{kim-2014-convolutional,bai2018empirical, kalchbrenner2016neural}. Convolutions are lightweight and fast and have many interesting use-cases, notably for lightweight classification. In the era when LSTMs were the workhorses of NLP applications, convolutions were positioned nicely on the pareto frontier of the compute-performance curve. They are fast and lightweight, and unlike Transformers, they do not suffer from quadratic complexity. Our work is also well-aligned with the resurgence of interest in convolutions where \citep{wu2019pay} showed that convolutions can outperform self-attention on several sequence transduction tasks. Moreover, the necessity of the self-attention inductive bias in transformers have been also a subject of recent interest. Synthesizer models \citep{tay2020synthesizer} showed that transformers can still do pretty well without token-token dot product self-attention and a random attention matrix can perform competitively on certain tasks.

\section{Pre-Trained Convolution Models}
This section describes the pre-trained Convolution Model. For most of our experiments, we adopt depthwise separable convolutions \cite{kaiser2017depthwise,sifre2014rigid,chollet2017xception} which have shown to be fast and efficient variants of the standard convolution.

\subsection{Lightweight Depthwise Convolution}
This section introduces Lightweight Depthwise Convolutions \cite{wu2019pay} which forms the backbone of our pre-trained convolution model.
\subsubsection{Depthwise convolutions} Depthwise convolutions convolve independently over every channel. Given an input tensor $X$ of dimensions $n \times d$, the depthwise convolution, $D(X, W_{c,:},i,c)$ is defined as:
\begin{equation}
O_{i,c} = \sum^{k}_{j-1} W_{c,j} \cdot X_{i + j - \lceil \frac{k+1}{2} \rceil)},c    
\end{equation}
where $W \in \mathbb{R}^{d \times k}$ are the learnable parameters of the layer. $O_{i,c}$ is the output at position $i$ and channel $c$. The overall output is a tensor of $n \times d$ of identical shape as the input.
\subsubsection{Lightweight Convolutions} $L(.)$ are depthwise separable convolutions with (1) softmax-normalized kernels and (2) shared output channels and weight tying. Specifically, this is written as:
\begin{equation}
O^L_{i,c} =  \sum^{k}_{j-1} \text{softmax}(W_{\hat{c},j}) \cdot X_{i + j - \lceil \frac{k+1}{2} \rceil)},\hat{c}     
\end{equation}
where $\hat{c} = \frac{cH}{d}$. In short, parameters are shared every $\frac{d}{H}$ output channels. When $H=1$, this is equivalent to sharing all the weights of all channels.

\subsubsection{Dynamic Convolutions}
Dynamic Convolutions $D_Y(.)$ are a new form of lightweight convolutions introduced by \cite{wu2019pay}. The key idea is to learn position-specific kernels for performing lightweight convolutions. This can be written as:
\begin{align}
D_{Y} = L(X, f(X_i)_{h,:}, i, c),
\end{align}
where $f(.)$ is a linear transformation with parameters $W^Q \in \mathbb{R}^{H \times k \times d}$ that learns a position dependent kernel.

\subsection{Span-based Seq2Seq pre-training}
We adopt span-based sequence-to-sequence pre-training as per \cite{raffel2019exploring}. Specifically, given an input sequence, we randomly mask spans of lengths $L$ and replace them with a special sentinel token. The pre-training task is then to generate the masked tokens as targets. For example: \textit{Inputs: The happy cat sat [mask].} and \textit{Outputs: on the mat.}

\subsubsection{Convolutional Seq2Seq Architecture}
We implement a Seq2Seq \cite{sutskever2014sequence} architecture similar to \cite{wu2019pay}. The key difference when compared with Transformer architectures is that we replace the multi-headed self-attention with convolutional blocks. Instead of query-key-value transforms, we use gated linear unit projections following \cite{wu2019pay}. Each convolution block be written as:
\begin{align*}
X^1 &= W^{I}X \odot \text{sigmoid}(W^{S}X), \\
X^2 &= \text{ConvBlock}(X^1), \\ 
X^3 &= W^O(X^2),
\end{align*}
where $W^I, W^S, W^O$ are trainable parameters. We experiment with simple lightweight convolutions, dynamic convolutions and dilated convolutions in our experiments. Following \cite{wu2019pay,gehring2017convolutional}, the encoder-decoder attention remains untouched. The convention follows the backbone Transformer model in which we wrap each submodule with layer normalization and residual connectors. Hence, each Conv block is written as:
\begin{align*}
X_{A} &= \text{LayerNorm}(\text{Conv}(X)) + X, \\ 
X_{B} &= \text{LayerNorm}(\text{FFN}(X_A) + X_A, 
\end{align*}
where Conv is any of the convolution models that we explore in our experiments. FFN(.) is a two layer feed-forward network with ReLU activations in the middle. 

\subsubsection{Optimization} The model optimizes the token-wise cross-entropy loss and is trained with teacher forcing. 
\begin{align*}
L = \sum^{L}_{t=1} \sum^n_{i=1} \log(\pi^t_i) + (1 - y^t_i)\log(1-\pi^t_i),
\end{align*}
where $\pi^t_{i}$ is the prediction of class $i$ at time step $t$ and $y^t_{i}$ is the ground truth label of the class $i$ at time step $t$.

\section{Research Questions and Discussion}
Before we delve into our experiments, we establish a set of research questions and agenda we hope this work aims to bring clarity to.
\begin{itemize}
    \item \textbf{RQ1}: Do convolutions benefit from pre-training as much as Transformers?
    \item \textbf{RQ2}: Are convolutional models, pre-trained or otherwise, competitive with Transformer models? When do they perform well?
    \item \textbf{RQ3}: What are the benefits (if any) of using pre-trained convolution models over pre-trained Transformers? Are convolutions faster alternatives to self-attention based Transformers?
    \item \textbf{RQ4}: What are the failure modes, caveats and reasons to \textbf{not} use pre-trained convolutions?
     \item \textbf{RQ5}: Are certain convolution variants better than others?
\end{itemize}

\section{Experiments and Analysis} 
This section presents our analysis and results. 
\subsection{Datasets}
Our evaluation is based on the following datasets and tasks.

\begin{itemize}
\item \textbf{Toxicity Detection} - We use the \textsc{Civil Comments} \citep{DBLP:journals/corr/abs-1903-04561} and \textsc{Wiki Toxic Subtypes} dataset \citep{10.1145/3038912.3052591}. Given a piece of short text (originating from social media or wikipedia), the goal is to determine if the content is toxic, i.e., a binary classification task. For this task, we evaluate on both accuracy and F1 score.

\item \textbf{Sentiment Classification} - This is a binary classification task that determines the polarity of documents, sentences and/or tweets. We use the IMDb reviews dataset \citep{maas2011learning}, Stanford Sentiment Treebank (SST-2) \citep{socher2013recursive} dataset, along with Twitter Sentiment140 (S140) \citep{Sentiment140} dataset. 
\item \textbf{News Classification} - This is a task of topic categorization for news articles. We use the AGNews dataset \cite{zhang2015characterlevel}. This is a four-way classification task.

\item \textbf{Question Classification} We use the \textsc{TREC} fine-grained question classification dataset \cite{li-roth-2002-learning}. This task involves classifying questions into $46$ fine-grained question categories.

\item \textbf{Semantic Parsing / Compositional Generalization} Compositional generalization is the ability of models to generalize \textit{compositionally} outside of the training distribution. To be specific, it needs be able to handle unseen combinations at test time. For this task, we use the COGS dataset \cite{kim2020cogs}, a task of generating semantic representation of a given English sentence. For example, \textit{A cat smiled} $\rightarrow$ cat($x_1$) AND smile.agent($x_2,x_1$).



\end{itemize}
All of the datasets, with the exception of the recent COGS dataset \citep{kim2020cogs}, are Tensorflow datasets$\footnote{\url{https://www.tensorflow.org/datasets/catalog/overview}.}$.

For each dataset, we evaluate all models with and without pre-training (details in subsequent sections).  Table \ref{tab:stats} reports the statistics of the datasets used in this paper.
\begin{table}[ht]
    \centering
    \small
    \begin{tabular}{c|ccc}
    \hline
       Dataset / Task  & \# Train & \# Test & \# Class  \\
       \hline
       Civil Comments & 3,820,210& 205,781 & 2 \\
       Wiki Toxicity & 561,808 & 234,564& 2\\ 
       IMDb & 25,000 & 25,000 & 2 \\ 
       SST-2 & 67,000& 1,800 &2 \\ 
       S140 & 1,600,000 & 359 & 2 \\ 
       TREC &  4,500& 500&46\\ 
       AGNews &  120,000& 7,600  &4\\
       COGS & 24,000 & 3000 & N/A \\ 
         \hline
    \end{tabular}
    \caption{Statistics of datasets used in our experiments. Datasets are diverse in terms of domains, tasks and amount of labeled data.}
    \label{tab:stats}
\end{table}

\subsection{Experimental Setup}
This section describes our experimental setup.
\subsubsection{Models} Our models are largely based on sequence to sequence models, a paradigm that has demonstrated great success made evident by models such as BART \cite{lewis2019bart} and T5\citep{raffel2019exploring}. We implement our models in Mesh Tensorflow (MTF) \citep{shazeer2018mesh}, a library for distributed and efficient parallel model training that has similar API to Tensorflow. We train models that are of base size, which corresponds to $12$ layers each in the encoder and decoder, along with $3072$ dimensions for the feed-forward layers, a model dimension of $768$ and a total of $12$ heads. Our Transformer models are largely based on T5        \citep{raffel2019exploring}, which is considered the current state-of-the-art Transformer model for NLP tasks and hence serves as a strong baseline. For the convolution models, our lightweight convolution and dynamic convolution models have a window size\footnote{We believe that tuning the hyperparameters of the convolution models can result in even better performance. However, we decided to keep these hyperparameters simple for the start.} of $7$ across all layers, the number of unique depth filters is $2$. For dilated models, we use a filter size of $[4,4,7,7,15,15,15,15,31,31,31]$ for our 12 layer convolution model.

\subsubsection{Pre-training} We pre-train both our convolutional and Transformer models for 524K steps with a batch size of $128$. Given the input sequence length of $512$, this corresponds to $65536$ tokens per batch. For pre-training, we use the Colossal Cleaned CommonCrawl Corpus (C4) \citep{raffel2019exploring} dataset which has demonstrated impressive results on downstream tasks. We use the span based seq2seq objective as the pre-training objective as mentioned in earlier sections. The span size is set to $3$ and a corruption rate of $15\%$ is adopted. 
We use the Adafactor optimizer \citep{shazeer2018adafactor} with an inverse square root learning rate scheduler. Each pre-training run is performed using 16 TPU-v3 chips and takes approximately 12 hours to complete for models of base size.  

\subsubsection{Downstream Fine-tuning} We fine-tune the pre-trained models using the following set of hyperparameters: We use a constant learning rate which is tuned amongst $\{0.001, 0.0005, 0.0001\}$. The batch size is generally set to $64$ but occasionally set to $32$ for smaller datasets. Intuitively, sequence length is task dependent but generally approximately the 90th percentile for each task. We fine-tune for a maximum of $100K$ steps and report peak validation performance. Fine-tuning uses the same Adafactor optimizer as during training. We perform fine-tuning on similar hardware, i.e., typically 16 TPUv3 chips are used per fine-tuning job.

\subsection{Experimental Results}
This section describes our experimental setup and results.
\subsection{Results on Toxicity Detection}
Table \ref{tab:all_results} reports results on toxicity detection. On both toxicity detection datasets the pre-trained and no-pre-training (raw) setup, the best models are the dilated convolution models and the dynamic convolution models. In fact, all convolutional models outperform Transformers on both CivilComments and WikiToxic. Before pre-training, convolutions outperform Transformers by approximately $1.5$ absolute percentage points. The gap narrows after pre-training where Transformers see a better gain (e.g., $+5.1\%$ against $+4.3\%$) from pre-training over convolutions on the CivilComments dataset. However, the converse is true on WikiToxic - the only case of performance degradation after pre-training. Overall, on this task, convolutions are competitive to Transformers and outperform them.

\begin{table*}[]
    \centering
    \begin{tabular}{l|ccccccccc}
    \hline
    & \multicolumn{2}{c}{\textsc{CivilComment}} & \multicolumn{2}{c}{\textsc{WikiToxic}} & IMDb & SST-2 & S140 & TREC & News\\
    Model & Acc & F1 & Acc & F1 & Acc & Acc & Acc & Acc & Acc\\
    \hline
    \multicolumn{10}{c}{\textbf{No pre-training}} \\ 
    Trans.  & 77.22& 85.09 & 91.93& 95.45 & 84.81&  78.44 & 58.84 & 78.00 & 84.25\\ 
    Light  & 78.58 & 85.82 & 91.05 & 94.65 & \textbf{85.88} & 81.65 & 60.64 & \textbf{82.20} & \textbf{87.22} \\
    Dilat.  & \textbf{79.94} & \textbf{86.50} & \textbf{92.29} & 94.91 & 85.84 & 79.01 & 55.62 & 79.60 & 81.24 \\
    Dyna.  & 78.49 & 84.71 & 90.06& \textbf{95.66} & 85.69 & \textbf{82.80} & \textbf{60.84} & 80.20 & 85.13\\
    \hline 
    \multicolumn{10}{c}{\textbf{With pre-training}} \\ 
         Trans.  & 81.16 & 86.56 & 91.46& 95.12  & \textbf{94.16} & 92.09 & 61.65 & 93.60 & 93.54\\
         Light  & 81.47& 87.58 & 93.61& 96.48  & 93.60 & \textbf{92.20} &61.65 &93.60 & 93.63\\ 
         Dilat.  & 81.67 & \textbf{87.78} & \textbf{93.84}& 96.21 &  93.92 & 92.09 & \textbf{62.85} & \textbf{94.20} & 93.26 \\
         Dyna. & \textbf{81.83} &	87.71&93.76 & \textbf{96.53}& 93.35 & 91.59 & 62.45 & 92.40  & \textbf{93.93}\\
         \hline
          \multicolumn{10}{c}{\textbf{Gain  from pre-training}} \\ 
          Trans. & +5.1\% & +1.7\% & -0.6\% & -0.4\% & +11.0\% & +17.4\% & +4.7\% & +20.0\% & +11.0\%\\ 
          Light & +3.7\% & +2.1\% & +2.8\% & +1.9\% & +9.0\%  & +13.0\% & +1.7\% & +14.0\% & +7.3\%\\
          Dilat. & +2.1\% & +1.5\% & +1.7\% & +1.4\% & +9.4\% & +17.0\% & +13.0\% & +18.0\% & +14.8\%\\ 
          Dyn. & +4.3\% & +3.5\%& +4.1\% & +1.0\%  & +8.9\% & +10.6\% & +2.6\% & +15.2\% & +10.4\%\\
          \hline
    \end{tabular}
    \caption{Comparison of pre-trained Convolutions and pre-trained Transformers on toxicity detection, sentiment classification, question classification and news classification. All models have approximately $230M$ parameters and are 12 layered seq2seq architectures. Our findings show that convolutions (1) also benefit from pretraining and (2) are consistently competitive to transformer models with and without pretraining. }
    \label{tab:all_results}
\end{table*}

\subsection{Results on Sentiment Classification}

Results on Sentiment Classification (IMDb, SST-2 and S140) can be found in Table \ref{tab:all_results}. On the IMDb reviews dataset, the best non-pre-trained model is the lightweight convolution model, outperforming the Transformer model. The best pre-trained model is the Transformer model. However, all convolutional models come in close with less than a percentage point gap difference with pre-trained Transformers. On the SST-2 and S140 tasks, we observe that the best models are convolution-based, regardless of whether the model is pre-trained or not. 

\subsection{Results on Question Classification}
The best non-pre-trained model is the Lightweight Convolution model. For pre-trained models, convolutional models also outperform the pre-trained Transformer. On this task, while most models benefit significantly from pre-training, Transformers seem to benefit slightly more from pre-training.

\subsection{Results on News Classification}
Results on news classification seems to follow similar trends as other benchmarks. Convolutional models outperform Transformers both in non-pre-trained and pre-trained setups. The highest gain from pre-training is obtained from the dilated convolution model.

\subsection{Results on Compositional Generalization Challenge and Semantic Parsing}
We conduct additional experiments on semantic parsing and compositional generalization. The task is framed as a sequence generation task. We use the recently proposed \citep{kim2020cogs} dataset. On the in-distribution test set, Transformers and convolutions have identical performance $(95\%)$. On the generalization or out of distribution set, Transformers perform at $77.5\%$ while convolutions come in at $76.9$. While convolutions do not exactly outperform Transformers, they come in close enough to be considered competitive.

\subsection{Summary of Results}
On the seven tasks across a broad range of domains we find that (1) non-pre-trained convolutions are competitive and frequently outperform non-pre-trained Transformers, (2) pre-trained convolutions outperform pre-trained Transformers on six out of seven tasks. This answers \textbf{RQ2}. 

We also find that convolutions are able to benefit from pre-training, in a similar fashion to self-attention-based models. Hence, the benefits achieved by pre-training are not exclusive to Transformer models. This answers \textbf{RQ1}.  

Amongst the pre-trained convolutional models, we find that dilated convolutions and dynamic convolutions are generally better than lightweight convolutions, thus answering \textbf{RQ5}. 

Finally, we observe that relative performance (i.e., rankings) do change with pre-training. This definitely shows that there is some kind of effect from composing architectures with pre-training. The direct implication of this effect is that a model that performs well (relatively) without pre-training will not necessarily perform the best when pre-trained (and vice versa). Hence, aside from conflating architectures with pre-training schemes, we do also need to take note that different architectures may behave differently under pre-training.

\section{Discussion and Analysis}
This section expands on the results via a detailed analysis and discussion. We discuss the pros/cons of pretrained convolutions, the impact of pre-training on performance and also recommendations to the broader community. 

\subsection{When do we expect pre-trained convolutions to fail?}
In our experimental section, we observed the potential upsides of convolutional models over well-established pre-trained Transformers and observe that we are able to get quality improvements in certain cases. However, it might be good to further understand the drawbacks of convolutions. 

One obvious weakness of pre-trained convolutions are their \textbf{lack of cross-attention} inductive bias that comes for \textit{free} with self-attention in the Transformer encoder. For this reason, it is not a good idea to use pre-trained convolutions for tasks that requires modeling the relationship between two or more sequences. To verify this, we run experiments on SQuAD and MultiNLI and find that convolutions do not come close to Transformers just because of this missing inductive bias. This should be clearly distinguished when examining and evaluating models, as how the early SNLI leaderboard\footnote{\url{https://nlp.stanford.edu/projects/snli/}} distinguished between models that used cross-attention and models that did not.

Our initial evaluations on benchmarks like SQuAD/MNLI \citep{rajpurkar2016squad,williams2017broad} showed that pre-trained convolutions are indeed significantly lackluster. For example, convolutions only achieve $\approx 75\%$ accuracy on MultiNLI, while transformers easily achieve $\approx 84\%$ accuracy. Likewise, while transformers achieve about $\approx 90\%$ F1 on SQuAd, convolutions come in around $\approx 70\%$. This is entirely expected because there is no way the premise/question can interact with the hypothesis/context. (\textbf{RQ4}). However, our experiments show that this was only because they lack this cross-attention property. When we augment convolutions with a single layer of cross attention at the encoder, we find that pre-trained convolutions come close (a delta of $(\approx 1\%)$) to pre-trained Transformers on datasets such as MultiNLI \citep{williams2017broad}, achieving about $\approx 83\%$ accuracy. 

That said, we leave it to the practitioner to decide whether the cross-attention inductive bias is actually important for the problem at hand. We also like to emphasize that the pattern of concatenating sentence pairs is not necessary practical when scaling up since this requires inference on every permutation of sentence pairs. For this reason, dual encoder setups that do fast embedding space look-ups are more practical and feasible in practice \citep{avq_2020}. Given the strong performance of convolutions in a series of encoding tasks, we can expect pre-trained convolutions to do well in a dual encoder setup. 

\subsection{What are the benefits of pre-trained convolutions over Transformers?}
We observed a reasonable quality improvement from using convolutions over Transformers. This section discusses the additional benefit. 
\begin{figure}[ht]
    \centering
    \includegraphics[width=1.0\linewidth]{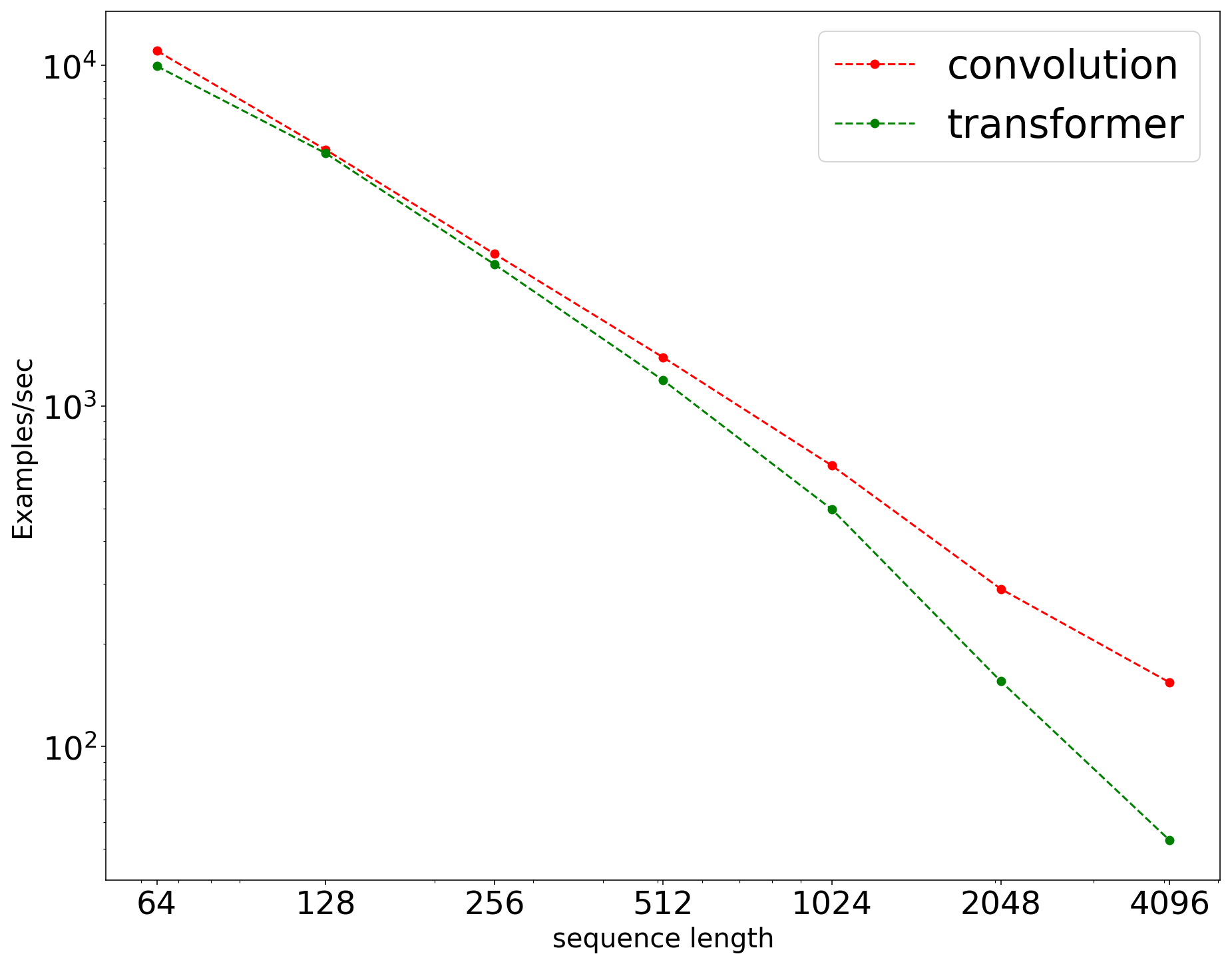}
    \caption{Effect of sequence length on processing speed (examples per second) on a seq2seq masked language modeling task. Results are benchmarked on 16 TPUv3 chips on C4 pre-training. Results are in log scale.}
    \label{fig:epsconv}
\end{figure}
\subsubsection{Convolutions are faster and scale better to long sequences}
Figure \ref{fig:epsconv} reports training speed of convolution (LightConvs) versus transformers on a sequence to sequence task. The input lengths are varied from $\{64,128,256,512,1024,2048,4096\}$. We show that convolutions are not only consistently faster (even at shorter sequences) but scale better than transformers. Convolution scales linearly while transformers are not able to scale to longer sequences.


\subsubsection{Convolutions are FLOPs efficient}
\begin{figure}[ht]
    \centering
    \includegraphics[width=1.0\linewidth]{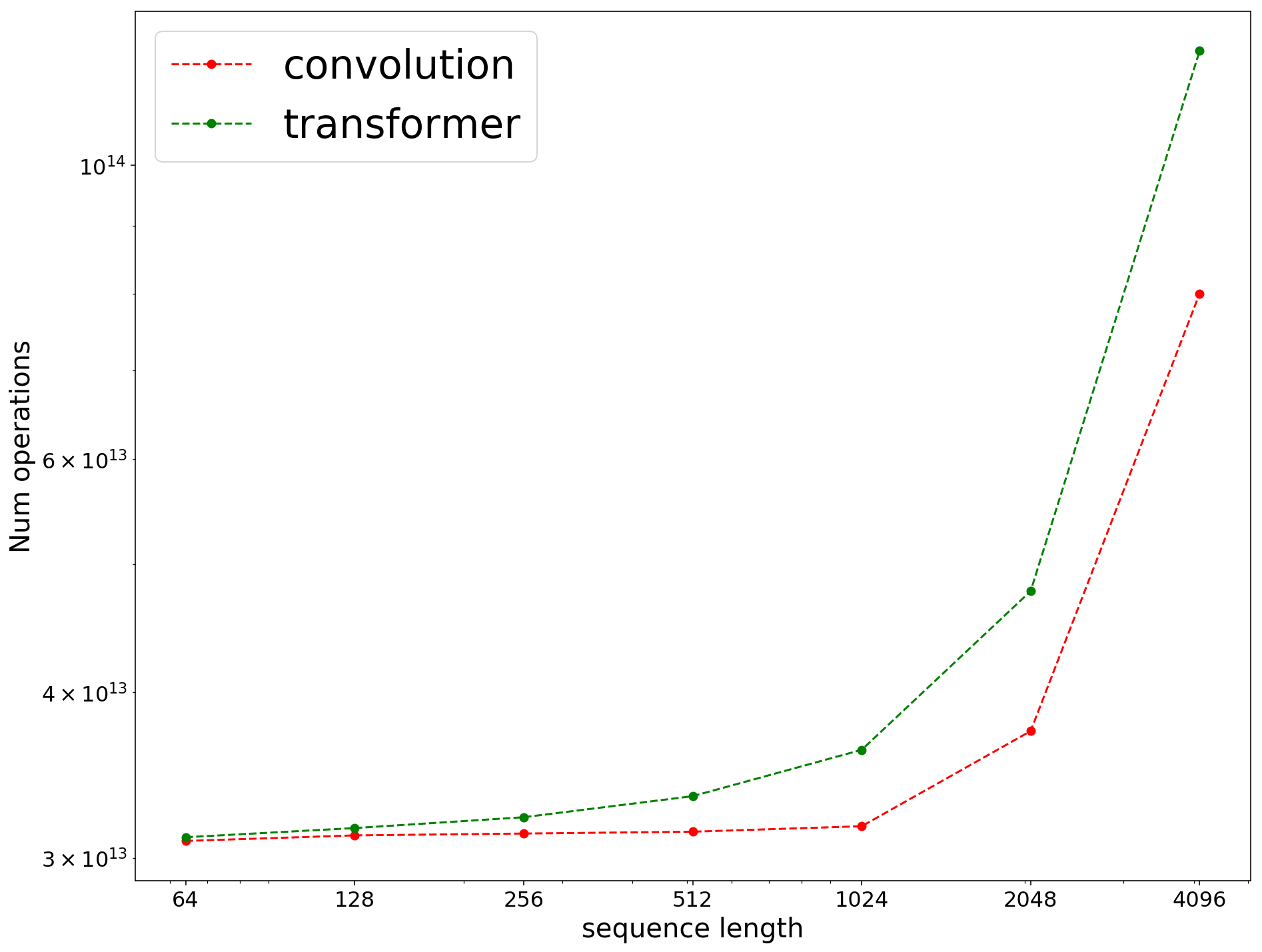}
    \caption{Effect of sequence length on number of FLOPs (einsum ops) on a seq2seq masked language modeling task. Results are benchmarked on 16 TPUv3 chips on C4 pre-training. Results are in log scale.}
    \label{fig:flopsconv}
\end{figure}

We measure the number of FLOPs of convolutions versus transformers as we increase the sequence length. Figure \ref{fig:flopsconv} shows the phenomenon while varying sequence length. In general, across all sequence lengths, convolutions are more efficient in the number of floating point operations.

The overall findings that convolutions are faster both in wall clock time and in FLOPs answers \textbf{RQ3}. Moreover, we find that the FLOP efficiency of convolutions scales better across sequence lengths.


\subsection{Are we suggesting to completely replace Transformers with convolution?}
While Transformers have dominated the research landscape in NLP, this paper suggests that there are commonly overlooked benefits to convolutions such as model quality, speed, FLOPs and scalability. Moreover, it is previously unknown to whether convolutions benefit from pre-training. In this paper, we showed that they are competitive on \textbf{some} tasks and also benefit from pre-training in similar fashion to transformer models. However, on the flip side, we also highlighted that they are \textbf{unable} to handle tasks that require cross-attention or when there is a need to model $>1$ sentence or documents within the same sequence. We believe that practitioners have good options and it might be worthwhile to explore architectures outside the well-established transformer models.

\subsection{On not conflating pre-training with architectural advances}
In this paper, we showed that three other (convolutional-based) architectures (e.g., lightweight, dymamic and dilated) also benefit from pre-training to the same extent as transformer models. 

In the current research landscape, pre-training has always be tightly coupled and associated with transformers architectures. As a result, the success of BERT, transformers and large language models seem to be pretty conflated. While it is true that, to this date, the only model that large-scale pre-training has been applied to are transformer models, we believe there might be potential in other architectures. 

Based on our empirical findings, we believe there is still significant room for the improving the understanding of the compositional effects of architecture and pre-training. Hence, we believe that the impact of this work extends beyond showing the competitiveness of convolution models in NLP. More concretely, the take home message is that there should be a healthy level of optimism in exploring architectural alternatives.


\section{Conclusion}
In this paper, we conducted an extensive study of the viability and feasibility of pre-trained convolutions. Our experimental results show that convolutions can outperform Transformers in \textbf{both} pre-train and non-pre-trained setups. Our extensive experiments across 8 datasets spanning a diverse range of tasks, show that convolutions are able to benefit from pre-training to the same (or sometimes greater) extent than Transformers. While pre-trained transformers are the de-facto choice of architecture, our results show that they might not be the best in certain scenarios. Additionally, we discussed the caveats, trade-offs pertaining with runtime, scalability, number of FLOPS and model quality. Finally, we discussed the situations or data types that convolutions are not well equipped to handle and make an empirically informed recommendation for practitioners.

\bibliography{ref}
\bibliographystyle{acl_natbib}

\appendix

\end{document}